%% file: main.tex
\documentclass[10pt,twocolumn,letterpaper]{article}

\usepackage{wacv}
\usepackage{times}
\usepackage{epsfig}
\usepackage{graphicx}
\usepackage{amsmath}
\usepackage{amssymb}
\usepackage{booktabs}

\usepackage{multirow}
\usepackage{booktabs}
\usepackage{adjustbox}
\usepackage{colortbl}
\usepackage{float}
\usepackage{stfloats}
\usepackage[accsupp]{axessibility}  
\renewcommand{\paragraph}[1]{\vspace{0.5em}\noindent\textbf{#1}~~}

\def\wacvPaperID{61} 

\wacvapplicationstrack 

\wacvfinalcopy 

\ifwacvfinal
\usepackage[breaklinks=true,bookmarks=false]{hyperref}
\else
\usepackage[pagebackref=true,breaklinks=true,colorlinks,bookmarks=false]{hyperref}
\fi

\begin{document}

\title{Diffeomorphic Image Registration with Neural Velocity Field}

\author{Kun Han\textsuperscript{1}\;\;\; 
Shanlin Sun\textsuperscript{1}\;\;\; 
Xiangyi Yan\textsuperscript{1}\;\;\; 
Chenyu You\textsuperscript{2}\;\;\; 
Hao Tang\textsuperscript{1} \\
Junayed Naushad\textsuperscript{1}\;\;\; 
Haoyu Ma\textsuperscript{1}\;\;\; 
Deying Kong\textsuperscript{1}\;\;\; 
Xiaohui Xie\textsuperscript{1}\\
\textsuperscript{1}University of California, Irvine, USA \;\;\;\;\;\; \textsuperscript{2} Yale University, USA\\
{\tt\small \{khan7,shanlins,xiangyy4,htang6,jnaushad,haoyum3,deyingk,xhx\}@uci.edu} \\
{\tt\small \{chenyu.you\}@yale.edu}
}

\maketitle

\thispagestyle{empty}

\begin{abstract}
Diffeomorphic image registration, offering smooth transformation and topology preservation, is required in many medical image analysis tasks.Traditional methods impose certain modeling constraints on the space of admissible transformations and use optimization to find the optimal transformation between two images. Specifying the right space of admissible transformations is challenging: the registration quality can be poor if the space is too restrictive, while the optimization can be hard to solve if the space is too general. Recent learning-based methods, utilizing deep neural networks to learn the transformation directly, achieve fast inference, but face challenges in accuracy due to the difficulties in capturing the small local deformations and generalization ability. Here we propose a new optimization-based method named DNVF (\underline{D}iffeomorphic Image Registration with \underline{N}eural \underline{V}elocity \underline{F}ield) which utilizes deep neural network to model the space of admissible transformations. A multilayer perceptron (MLP) with sinusoidal activation function is used to represent the continuous velocity field and assigns a velocity vector to every point in space, providing the flexibility of modeling complex deformations as well as the convenience of optimization. Moreover, we  propose a cascaded image registration framework (Cas-DNVF) by combining the benefits of both optimization and learning based methods, where a fully convolutional neural network (FCN) is trained to predict the initial deformation, followed by DNVF for further refinement. Experiments on two large-scale 3D MR brain scan datasets demonstrate that our proposed methods significantly outperform the state-of-the-art registration methods.
\end{abstract}

\section{Introduction}


Image registration is an essential  task  used in many medical image analysis applications \cite{incoronato2017radiogenomic,risholm2011multimodal}, such as assessing disease progression over time, merging and comparing different image modalities, and shape analysis. By maximizing the image similarity, such as intensity correlation, image registration provides the correspondence and non-linear transformation between pairs of images. Diffeomorphic image registration offers more desirable properties such as smooth deformation, topology preservation, and transformation invertibility.

Traditional methods, such as elastic-type models~\cite{bajcsy1989multiresolution,shen2002hammer}, B splines~\cite{rueckert1999nonrigid}, LDDMM~\cite{beg2005computing,zhang2017frequency} and SyN~\cite{avants2008symmetric}, solve the image registration problem by optimizing the deformation fields. These methods typically make certain model assumptions. For example, LDDMM assumes the diffeomorphic deformation can be obtained by solving the flow-based ordinary differential equation (ODE) with certain regularization constraints. SyN symmetrizes the cross-correlation Euler-Lagrange equations within the space of diffeomorphic maps. However, these methods usually generate high-accuracy results at the cost of slow speed and intensive computation, and the performance may vary under different modeling assumptions \cite{sotiras2013deformable}.

Rapid advance in learning-based methods~\cite{balakrishnan2018unsupervised,dalca2018unsupervised,sheikhjafari2018unsupervised,mok2020large,mok2020fast,zhang2021learning,shen2021accurate,mok2021conditional} have achieved promising results in the image registration task. With deep neural networks, learning-based methods can efficiently estimate the transformation between two medical images. As more attention is focused on learning-based methods, many new techniques and complex network structures \cite{chen2021vit,mok2022affine,hu2021end,shi2022xmorpher} have been applied to chase better performance. However, the accuracy improvement is only modest, mainly because the representations learned from neural networks are not able to predict sophisticated deformations and dense correspondences for each pair of images in dataset. Moreover, the generalizability is still a major challenge for these methods which limits the performance for the out-of-distribution image pairs. 

The recent development of neural fields provides a class of coordinate-based neural networks which parameterize the physical properties of objects across space and time \cite{xie2022neural}. Neural fields have shown their great potential in modeling general dynamic scenes \cite{pumarola2021d,li2021neural,sun2022topology} which fit the observed time-variant views with great detail through the optimization of a neural network. Therefore, our question of curiosity is hence: \textit{can we use neural fields to represent the dynamics of diffeomorphic image registration?}

In this paper, we propose to realize diffeomorphic image registration by optimizing an implicit neural representation of a continuous velocity field. Specifically, we parameterize the continuous velocity field as a multilayer perceptron (MLP), whose input is a 3D spatial coordinate ($x,y,z$) and output is the corresponding 3D velocity vector ($v_x,v_y,v_z$).  With periodic sinusoidal activation functions, the MLP can efficiently represent the high-frequency content \cite{sitzmann2020implicit} and therefore improve its ability to model the small and complex deformations in the registration problem. The diffeomorphic deformation can be obtained through integration over the neural velocity field, which is realized by the scaling and squaring (SS) method \cite{arsigny2006log} in our work. SS follows the Lie algebra in group theory and the deformation is produced by the exponential of the velocity field through a spatial transform layer.

Moreover, we propose a cascaded framework called Cas-DNVF which combines the benefits of learning-based methods and DNVF. In the first stage, we pretrain a fully convolutional neural network to predict an initial deformation with a short inference time and simplify the search space of optimal deformation for the following DNVF. Based on that, DNVF can optimize a residual deformation specifically for each pair of images. By combining the benefits of two different methods, Cas-DNVF has better generalizability and can achieve the accurate alignment between two images within a short running time. Our experiments shows the DNVF can be integrated with different learning-based registration methods under the framework of Cas-DNVF.

Our contributions of this work are as follows:

\begin{itemize}
    \item We propose neural registration method, called DNVF, for diffeomorphic image registration, utilizing MLP with sinusoidal activation to represent continuous velocity field and model diffeomorphic deformation through integral curves of velocity field. Optimal registration is discovered by tuning parameters of MLP. 
    \item We further propose a cascaded framework (Cas-DNVF) to incorporate learning to DNVF, where a fully convolutional net is trained to predict the initial deformation, followed by DNVF for further refinement. 
    \item Extensive experiments on two 3D brain MR datasets demonstrate that the proposed methods achieve state-of-the-art performance while preserving desirable diffeomorphic properties.
\end{itemize}

\section{Related works}

\subsection{Pair-wise optimization method}

\vspace{-0.5em}

Extensive works have been conducted to tackle the task of image registration. Traditional methods model image registration as an optimization problem and minimize the energy function iteratively for each pair of images \cite{sotiras2013deformable}. These methods typically enforce transformation regularity through certain model assumptions. Several studies directly optimize the deformable displacement field including elastic-type models \cite{bajcsy1989multiresolution,shen2002hammer}, statistical parametric mapping \cite{ashburner2000voxel}, free-form deformations with b-splines \cite{rueckert1999nonrigid} and optical flow based Demons \cite{thirion1998image}. Besides, many other studies focus on the registration problem within the space of diffeomorphic maps to ensure the desirable diffeomorphic properties, such as topology preservation and transformation invertibility. Popular methods include
Large Deformation Diffeomorphic Metric Mapping (LDDMM) \cite{beg2005computing,zhang2017frequency} and symmetric image normalization method (SyN) \cite{avants2008symmetric}. LDDMM models the diffeomorphic deformation by considering the velocity over time according to the Lagrange transport equation \cite{christensen1996deformable,dupuis1998variational}. And SyN develops a novel symmetric diffeomorphic optimizer for maximizing the cross-correlation in the space of topology preserving maps \cite{avants2008symmetric}. 

The proposed DNVF is also a pair-wise optimization-based diffeomorphic image registration method, however, there are no strong assumptions about the dynamics of the registration since using a neural network to model the deformation provides greater flexibility. Moreover, DNVF can utilize deep learning packages for efficient inference and optimization. 

\subsection{Learning-based method}
\vspace{-0.5em}
Medical researches have shown the promising progress brought by the recent learning methods \cite{ronneberger2015u, chen2021transunet, chen2021deep,tang2021spatial,yan2022after,you2022class,tang2021recurrent,you2022incremental,you2022bootstrapping,you2022mine}. In image registration, learning-based methods \cite{balakrishnan2018unsupervised,dalca2018unsupervised,sheikhjafari2018unsupervised,shen2019networks,shen2019region,mok2021conditional,mok2020large,mok2020fast,zhang2021learning,chen2021vit,mok2022affine,hu2021end,shi2022xmorpher,zhang2022learning} achieve higher accuracy and efficiency. By learning a common representation for a collection of images, the extracted features can be used to perform registration with fast inference speed. VoxelMorph \cite{balakrishnan2018unsupervised} directly regresses deformation fields by minimizing the dissimilarity between input and target images. The multi-resolution strategy was introduced in LapIRN \cite{mok2020large} to avoid local minima during optimization. SYMNet \cite{mok2020fast} symmetrically warps images regarding the middle of the geodesic path and predicts the diffeomorphic deformation. A recursive cascaded network was proposed in \cite{zhao2019recursive} to boost the performance of registration by iteratively applying the registration network to the warped moving image and fixed image. A transformer block was deployed over the CNN backbone to capture the semantic contextual relevance in DTN \cite{zhang2021learning}. To better solve large deformations, MS-ODENet \cite{xu2021multi} proposed to use neural ODE on image registration to refine the estimated transformation, by modeling the dynamics of the parameters of registration models. However, the representation power of neural networks is limited by the network structure and training data might not be able to generate complex deformations and capture dense correspondences for all pairs of images in a dataset. The generalization gap between training data and test data also restricts the performance of a pre-trained neural network during inference time.

\subsection{Neural Field}
\vspace{-0.5em}
The recent advance of neural fields enables the parameterization of physical properties and dynamics through coordinate-based neural networks. \cite{park2019deepsdf} formulated the generative shape-conditioned 3D modeling with a continuous implicit surface. \cite{sitzmann2020implicit} introduced sinusoidal representation networks to model the 2D image and 3D scene with fine details. \cite{niemeyer2019occupancy} learned a temporally and spatially continuous vector field to perform dense 4D reconstruction from images or sparse point clouds. \cite{wu2021irem} proposed to perform high-resolution MR image reconstruction via implicit neural representation. \cite{sun2022topology} applied neural field to model the diffeomorphic transformation on 3D shapes. Some recent works focus on pair-wise image registration problems, such as IDIR \cite{wolterink2021implicit} and NODEO \cite{wu2022nodeo}. 

Unlike them, DNVF uses a simple multilayer perceptron to represent the continuous neural velocity field and model the diffeomorphic deformation. The proposed Cas-DNVF further combines the benefit of learning-based and optimization-based methods with better generalizability, matching accuracy and time efficiency. Experimental results in Section \ref{sec:exp} demonstrate the advantages of our proposed methods.

\section{Preliminaries}

\subsection{Deformable registration}
\vspace{-0.5em}
Deformable image registration denotes warping one (moving) image to align it with the second (fixed or target) image by maximizing the similarity between the registered images under some regularization constraints. The displacement field returned from the registration defines the dense mapping between points in the moving image and corresponding points in the fixed image. The typical deformable image registration can be formulated as : 
\begin{equation}
    \phi^{*}=\underset{\phi}{\arg \min }\ \mathcal{L}_{\text {sim }}(I_f, \phi \circ I_m)+\mathcal{L}_{r e g}(\phi)
\end{equation}
where $\phi^{*}$ represents the optimal displacement field $\phi$, $I_f$ and $I_m$ denote the fixed and moving images, $\phi \circ I_m$ represents $I_m$ warped by $\phi$, $\mathcal{L}_{s i m}$ measures the image similarity between the fixed image and warped image, and $\mathcal{L}_{r e g}$ represents the smoothness regularization function.


\subsection{Diffeomorphic registration}
\label{sec:3.2}
\vspace{-0.5em}
Diffeomorphic image registration not only aligns two images but also preserves the topology and maintains transformation invertibility \cite{beg2005computing}. The diffeomorphic deformation $\phi$ is calculated through the integral of the velocity field $\boldsymbol{v}$ (assume Lipschitz continuity) following the ordinary differential equation (ODE):
\begin{equation}
    \frac{\partial \boldsymbol{\phi}^{(t)}}{\partial t}=\boldsymbol{v}(\boldsymbol{\phi}^{(t)})
\end{equation}
where $\boldsymbol{\phi}^{(0)}=I$ is the identity transformation. In this paper, we assume the velocity  field $\boldsymbol{v}$ is stationary over $t=[0,1]$ and the final deformation is taken to be $\boldsymbol{\phi}^{(1)}$.  

\section{Method}

\begin{figure*}[ht]
\begin{center}
\includegraphics[width=0.65\linewidth]{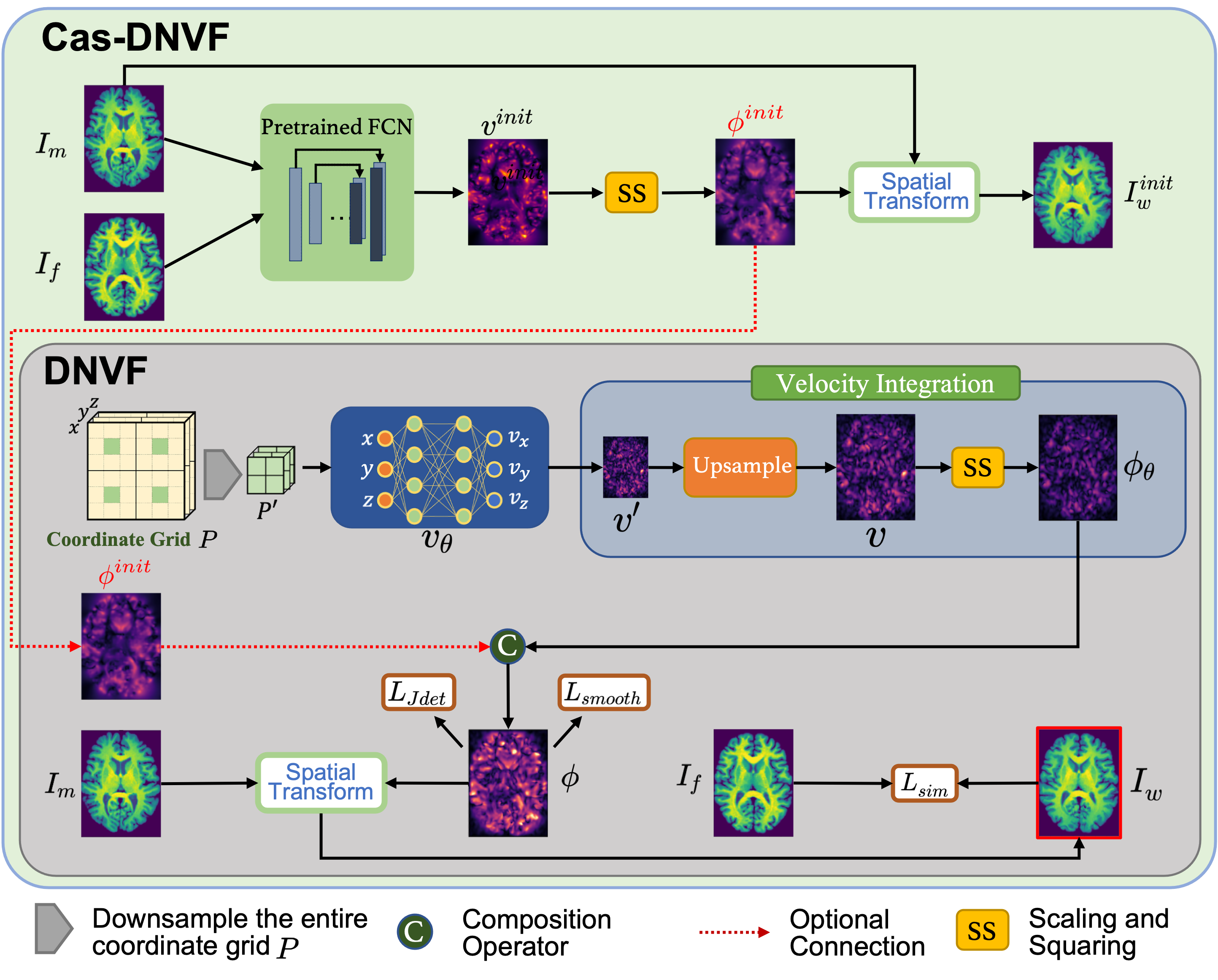}
\end{center}
\vspace{-1.5em}
   \caption{Overall framework of the proposed DNVF and Cas-DNVF. $I_m$, $I_f$ and $I_w$ denote the moving image, fixed image and  warped image. DNVF uses neural representations to model continuous velocity field $v_\theta$ which takes 3D position $\boldsymbol{p}\in\mathbb{R}^3$ as input and assigns corresponding velocity vector $\boldsymbol{v}\in\mathbb{R}^3$. The velocity field $v_\theta$ is represented by a MLP with periodic sinusoidal activation functions. The velocity integration inside DNVF will be introduced in Sec.\ref{sec:4.2}. During the optimization, DNVF implicitly captures the dense correspondence between two input images. Cas-DNVF combines learning-based method and DNVF. A FCN is firstly pretrained to predict the initial deformation and simplify the search space of optimal deformation for the following DNVF so that it can focus on modeling the small local deformation with high accuracy and efficiency. The framework is designed for 3D registration, we use 2D image for simplicity.}
\label{fig:model_all}
\vspace{-1.5em}
\end{figure*}

Let $I_{f}$, $I_{m}$ be the fixed image and moving image that need to be aligned. In this paper, we focus on 3D image registration where $I_{f}$ and $I_{m}$ are defined on 3D spatial domain $\Omega \subset \mathcal{R}^{3}$. $I_{f}$ and $I_{m}$ are affinely registered in the preprocessing step, therefore, we only need to model the non-linear displacement between two images. 

Figure \ref{fig:model_all} presents an overview of our methods. DNVF is an optimization-based model which utilizes a MLP to represent the neural velocity field $v_\theta$ where $\theta$ are the parameters of the MLP. Unlike previous image registration methods, DNVF takes the 3D spatial coordinates as the input, rather than the image intensities. For each spatial point $\boldsymbol{p} \in \Omega$, $v_\theta$ provides the corresponding velocity vector $\boldsymbol{v} = v_\theta(\boldsymbol{p})$ at that point. The diffeomorphic deformation $\phi_\theta$ is then calculated through the integral over the neural velocity field $v_\theta$ as described in Sec. \ref{sec:3.2}. Inside DNVF, we use scaling and squaring to do the integration and the details will be described in Sec. \ref{sec:4.2}. We optimize the parameters $\theta$ of the neural velocity field and find the optimal $\hat{\theta}$ by minimizing the loss function:
\begin{equation}
\begin{split}
    \hat{\theta}=\underset{\theta}{\arg \min }\ \mathcal{L}_{\text {sim }}(I_f, \phi_\theta \circ I_m )+\mathcal{L}_{r e g}(\phi_\theta)
\end{split}
\end{equation}

Based on DNVF, we propose a cascaded framework Cas-DNVF which combines the benefit of the learning-based methods and DNVF. First, we train a fully convolutional neural network (FCN) to model a function $g_{\beta}(I_f, I_m)=\phi^{init}$ which provides an initial deformation for a given pair of images. We train the $g_{\beta}$ by minimizing the loss function similar to Eq.3:
\begin{equation}
\begin{aligned}
    \hat{\beta}= \underset{\beta}{\arg \min }\ [\mathbb{E}_{(I_f, I_m) \sim D}[& \mathcal{L}_{\text {sim }}(I_f, g_{\beta}(I_f, I_m) \circ I_m)+\\
    & \mathcal{L}_{r e g}(g_{\beta}(I_f, I_m))]] 
\end{aligned}
\end{equation}
where $\mathcal{D}$ is the dataset distribution, $\beta$ are the parameters of the FCN, $I_f$ and $I_m$ are the sampled volume pairs from $\mathcal{D}$. During inference, for the input image pair ($I_f$, $I_m$), we first use the pretrained $g_{\beta}$ to predict an initial deformation $\phi^{init}$. In the second stage, we use DNVF to optimize a residual deformation $\phi_\theta^{res}$ where the overall deformation is calculated using a spatial transform layer to combine $\phi^{init}$ and $\phi_\theta^{res}$:
\begin{equation}
    \hat{\theta}=\underset{\theta}{\arg \min }\ \mathcal{L}_{\text {sim }}(I_f, \phi_{\theta}^{res} \circ \phi^{init} \circ I_m) +\mathcal{L}_{r e g}(\phi_\theta^{res} \circ \phi^{init})
\end{equation}
By analyzing the performance shown in Section \ref{sec:exp}, the initial deformation $\phi^{init}$ predicted by $g_\beta$ helps DNVF to achieve faster convergence while alleviating the generalizability issue of learning-based method and providing a more precise dense matching between the pair of input images.

\subsection{Neural Velocity Field Representation}
\label{sec:4.1}
\vspace{-0.5em}
Inspired by recent works on neural rendering \cite{sitzmann2020implicit,mildenhall2020nerf}, we model the neural representation of the continuous velocity field $v_\theta$ using a MLP where $\theta$ are the parameters of the neural network. The neural velocity field can be viewed as a function of a spatial 3D point: $\boldsymbol{v}=v_\theta(\boldsymbol{p})$, which outputs the corresponding velocity vector $\boldsymbol{v}$ given 3D spatial coordinate $\boldsymbol{p}$. 

In this paper, we focus on the diffeomorphic registration between two 3D brain MR scans. The complex structure of the human brain requires sophisticated deformation field $\phi_\theta$ to achieve a precise matching. Therefore, we expect the neural velocity field $v_\theta$ to be able to model the high frequency function, otherwise, the deformation $\phi_\theta$ integrated over $v_\theta$ will be too smooth and can not provide an accurate mapping. However, classic MLPs have difficulty learning high frequency functions because of the "spectral bias" \cite{tancik2020fourier}. As shown in \cite{rahaman2019spectral,basri2020frequency}, the deep networks have a learning bias towards low frequency functions. 

Therefore in this work, instead of using classic ReLU activation function, we choose to use periodic sinusoidal function enabling fitting of high-frequency content as indicated by SIREN \cite{sitzmann2020implicit} and adopt its weight initialization scheme for deep structure. The neural velocity field $v_\theta$ is designed as follows: 
\begin{equation}
\begin{aligned}
    & f_{0}(\mathbf{x}_{0}) = \mathbf{W}_{0} \mathbf{x}_{0}+\mathbf{b}_{0} \mapsto x_1\\
    & f_{i}(\mathbf{x}_{i}) =\mathbf{W}_{i}\sin ( \mathbf{x}_{i})+\mathbf{b}_{i} \mapsto x_{i+1}
\end{aligned}
\end{equation}
where the 3D coordinate $\boldsymbol{p}$ is firstly mapped to a high dimensional embedding by $f_{0}: \mathbb{R}^{3} \mapsto \mathbb{R}^{N}$. Then the $i^{th}$ layer of network $f_i$ can be viewed as a Fourier frequency mapping \cite{benbarka2022seeing} with the learnable parameters $\mathbf{W}_i$ and $\mathbf{b}_i$. By involving the frequency information in the network, the neural tangent kernel \cite{tancik2020fourier} is modified accordingly such that the neural velocity field $v_\theta$ can have a good representation of the high-frequency details and be able to model the sophisticated deformation $\phi_\theta$. In our implementation, we use a 5-layer MLP to represent the neural velocity field $v_\theta$ with 512 hidden units.

\subsection{Diffeomorphic Deformation as Integration}
\label{sec:4.2}
\vspace{-0.5em}
Inside DNVF, after defining the neural velocity field $v_\theta$, the diffeomorphic deformation $\phi_\theta$ is calculated by integrating over $v_\theta$ according to (2). Following \cite{dalca2018unsupervised,mok2020fast,mok2020large,zhang2021learning}, the velocity field is assumed to be stationary over time. We use the scaling and squaring method to calculate the diffeomorphic deformation as shown in Figure \ref{fig:model_all}:

\paragraph{Scaling and Squaring \cite{arsigny2006log}} When the velocity field is stationary, the exponential map $\boldsymbol{\phi}^{(t)}=\exp(v_\theta)$ defines one-parameter subgroup of diffeomorphisms.  The final deformation $\boldsymbol{\phi}^{(1)}=\exp(v_\theta)$ can be solved more efficiently by utilizing the group actions.  Specifically, the initial deformation is $\phi^{(1 / 2^{T})}=\boldsymbol{p}+v_\theta(\boldsymbol{p})/2^{T}$ where $T$ is the total time step. We recursively compute $\phi^{(1 / 2^{t-1})}=\phi^{(1 / 2^{t})} \circ \phi^{(1 / 2^{t})}$ through a spatial transform layer, and the final deformation $\phi^{(1)}$ is obtained by $\phi^{(1)}=\phi^{(1/2)} \circ \phi^{(1 / 2)}$. No additional learnable parameters are introduced during the recursive operation. The deformations are calculated over a fixed grid with linear interpolation. The entire step is differentiable. In our implementation, we chose $T=7$.

It is infeasible to feed all 3D coordinates ($D \times H \times W \times 3$) into DNVF due to GPU memory constraints. Therefore in our implementation, we empirically downsample the original coordinate grid with a scale 1/3 and upsample the resulting velocity field to recover the full resolution for velocity integration as shown in Figure \ref{fig:model_all}.

\subsection{Cascaded Registration}
\begin{figure}[ht]
\begin{center}
\includegraphics[width=\linewidth]{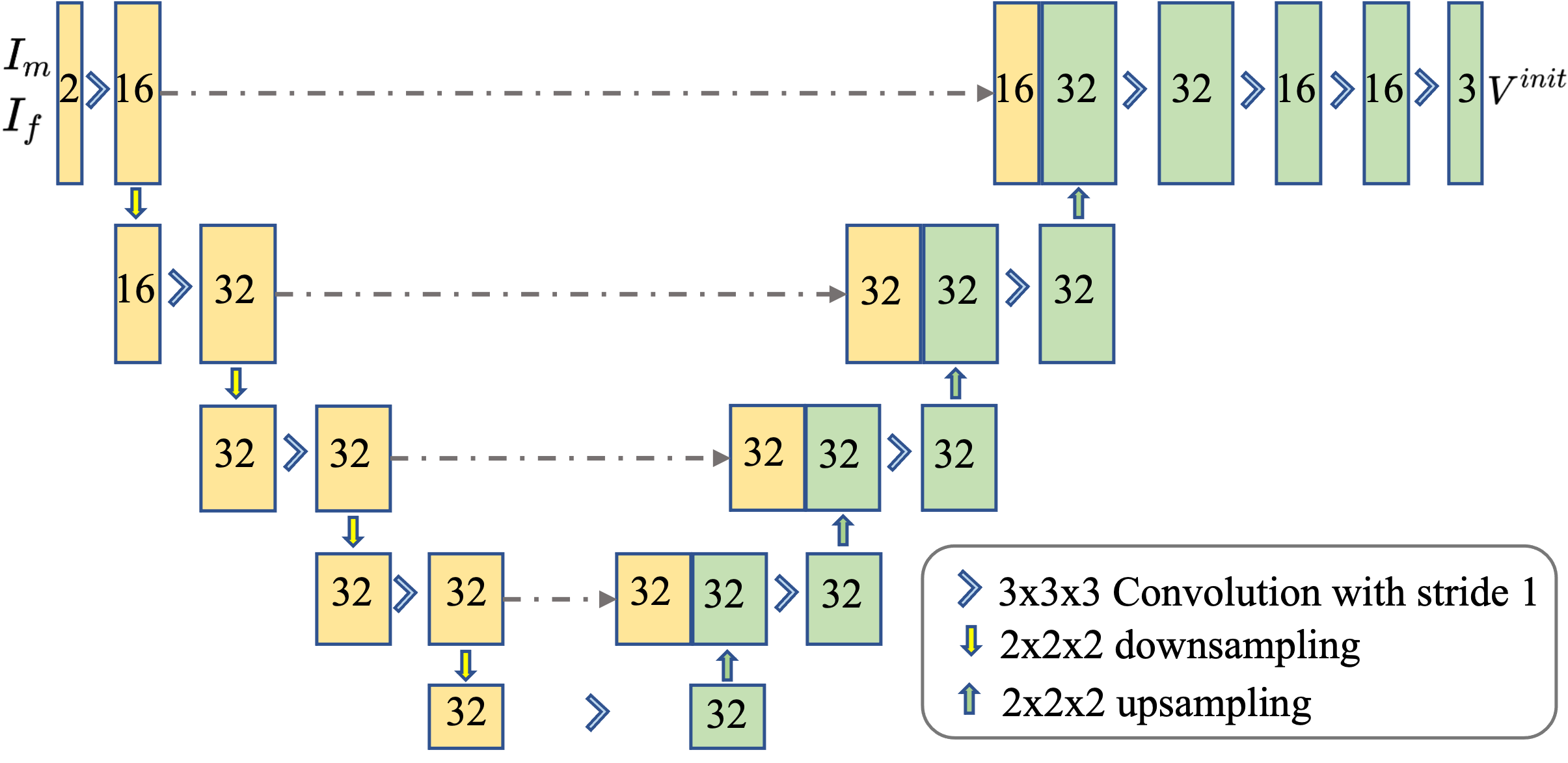}
\end{center}
\vspace{-0.5em}
   \caption{Structure of fully convolutional network (FCN) used in Cas-DNVF. The input is the concatenation of moving image and fixed image, and the output is the velocity field. }
\label{fig:model_fcn}
\vspace{-1em}
\end{figure}

\vspace{-0.5em}
\paragraph{Initial Deformation} As shown in Figure \ref{fig:model_all}, we parameterize the function $g_{\beta}(I_f, I_m)=\phi^{init}$ in (4) with a fully convolutional neural network (FCN), a scaling and squaring layer, and a spatial transform layer. FCN adopts the Unet-like network structure as shown in Figure \ref{fig:model_fcn} which takes the concatenation of moving image and fixed image as input, and directly outputs the velocity field. By maximizing the similarity between the warped image and fixed image, the FCN is trained to predict a initial deformation $\phi^{init}$.

Moreover, $g_{\beta}$ can also be parameterized by other learning-based models such as \cite{dalca2018unsupervised,mok2020fast,mok2020large,zhang2021learning}. Ablation study in Sec. \ref{sec:aba} shows that DNVF can provide consistent boost in performance for different learning-based methods.

\paragraph{Optimization of Residual Deformation} Based on the initial deformation predicted by the trained $g_\theta$, we use DNVF to optimize the residual deformation $\phi_\theta^{res}$ for each pair of images following (5). The $\phi^{init}$ and $\phi_\theta^{res}$ are combined by a spatial transform layer as the overall deformation $\phi$. 

Because the predicted $\phi^{init}$ usually provides a good mapping for large deformations, DNVF will focus more on the local small deformations which are difficult for learning-based methods. The velocity vector $\boldsymbol{v}$ output from $v_\theta$ accounts for the local small deformations between the warped moving image ($\phi^{init} \circ I_m$) and fixed image $I_f$. In our implementation, instead of directly combining two deformations, we empirically rescale the velocity vector $\boldsymbol{v}$ in DNVF by a factor 0.1 to improve the stability during optimization.

\subsection{Optimization}
\label{sec:opt}
\vspace{-0.5em}
The loss function $\mathcal{L}$ consists of two components: $\mathcal{L}_{sim}$ penalizing the misalignment between the warped moving image $I_w$ and fixed image $I_f$, $\mathcal{L}_{reg}$ regularizing the deformation smoothness ($\mathcal{L}_{smooth}$) and local orientation consistency ($\mathcal{L}_{Jdet}$). 

\paragraph{$\boldsymbol{\mathcal{L}_{sim}}$} We use local normalized cross-correlation (NCC) as the metric to measure the similarity between the warped moving image $I_w$ and fixed image $I_f$:
\begin{equation}
\begin{aligned}
    & NCC(I_w, I_f) = \\
    & \sum_{\boldsymbol{p} \in \Omega} \frac{\left(\sum_{\boldsymbol{p}_{i}}\left(I_w\left(\boldsymbol{p}_{i}\right)-\bar{I}_w(\boldsymbol{p})\right)\left(I_f\left(\boldsymbol{p}_{i}\right)-\bar{I}_f(\boldsymbol{p})\right)\right)^2}{\sum_{\boldsymbol{p}_{i}}\left(I_w\left(\boldsymbol{p}_{i}\right)-\bar{I}_w(\boldsymbol{p})\right)^{2} \sum_{\boldsymbol{p}_{i}}\left(I_f\left(\boldsymbol{p}_{i}\right)-\bar{I}_f(\boldsymbol{p})\right)^{2}}
\end{aligned}
\end{equation}
where $I_w = \phi\circ I_m$, $\phi$ is calculated by the velocity integration via SS, $\boldsymbol{p}_i$ denotes the points over a local window of size $n^3$ around $\boldsymbol{p}$, $\bar{I}_w(\boldsymbol{p})$ and $\bar{I}_f(\boldsymbol{p})$ are the mean intensity of that local window. High value of NCC represents a precise matching between images. Therefore, we use negative NCC as the similarity loss: $\mathcal{L}_{sim}=-NCC(\phi\circ I_m, I_f)$ with the window size set to 9.

\paragraph{$\boldsymbol{\mathcal{L}_{Jdet}}$} In order to secure local orientation consistency, we follow ~\cite{mok2020fast} to impose a selective Jacobian determinant regularization. If the Jacobian determinant at a given point $\boldsymbol{p}$ is positive, then the deformation field preserves the orientation near $\boldsymbol{p}$. Otherwise, the orientation in the neighborhood is reversed and the topology is destroyed.  With a ReLU function, we can penalize the local region with a negative Jacobian determinant:

\begin{equation}
    \mathcal{L}_{J d e t}=\frac{1}{N} \sum_{\boldsymbol{p} \in \Omega} relu\left(-\left|J_{\phi}(\boldsymbol{p})\right|\right)
\end{equation}
where the Jacobian matrix $J_\phi$ is defined as:
\begin{equation}
    J_{\phi}(\boldsymbol{p})=\begin{bmatrix}
\frac{\partial \phi_{x}(\boldsymbol{p})}{\partial x} & \frac{\partial \phi_{x}(\boldsymbol{p})}{\partial y} & \frac{\partial \phi_{x}(\boldsymbol{p})}{\partial z} \\
\frac{\partial \phi_{y}(\boldsymbol{p})}{\partial x} & \frac{\partial \phi_{y}(\boldsymbol{p})}{\partial y} & \frac{\partial \phi_{y}(\boldsymbol{p})}{\partial z} \\
\frac{\partial \phi_{z}(\boldsymbol{p})}{\partial x} & \frac{\partial \phi_{z}(\boldsymbol{p})}{\partial y} & \frac{\partial \phi_{z}(\boldsymbol{p})}{\partial z}
\end{bmatrix}
\end{equation}
\vspace{-0.5em}
\paragraph{$\boldsymbol{\mathcal{L}_{smooth}}$} In order to avoid oddly skewed deformations, a spatial gradient is used to constrain the smoothness of the deformation field as a regularization term. A large spatial gradient means the radical change of deformation in the local area which is not desired in the registration problem. Therefore, the smoothness loss is defined as: 
\begin{equation}
    \mathcal{L}_{smooth}= \sum_{\boldsymbol{p} \in \Omega}\|\nabla \phi(\boldsymbol{p})\|^{2}
\end{equation}

\vspace{-0.5em}

We present the complete loss function as follows, where $\lambda_{1}$ and $\lambda_{2}$ control the weight of orientation consistency loss and deformation smoothness loss:
\begin{equation}
    \mathcal{L}=\mathcal{L}_{sim}+\mathcal{L}_{reg} = \mathcal{L}_{sim}+\lambda_{1}\mathcal{L}_{Jdet}+\lambda_{2}\mathcal{L}_{smooth}
\end{equation}

\section{Experiment}
\label{sec:exp}
\subsection{Dataset and Preprocessing}
\vspace{-0.5em}
We evaluate our method on two public 3D brain MR datasets: the OASIS \cite{marcus2007open} and the Mindboggle101 \cite{klein2012101}. \textbf{OASIS} dataset contains 416 T1-weighted MR scans aging from 18 to 96 with 100 of them diagnosed with mild to moderate Alzheimer's disease. Subcortical segmentation maps of 35 anatomical structures serve as the ground truth for the evaluation of our method. \textbf{Mindboggle101} consists of 101 T1-weighted MR scans from 5 datasets, e.g. HLN-12, MMRR-21 and NKI-RS. We followed \cite{xu2019deepatlas} to remove the images with incorrect labels and evaluated the performance on 31 cortical regions. Standard preprocessing methods were carried out on two datasets. Having skull stripped, all scans were resampled to same resolution $(1mm\times1mm\times1mm)$. For each dataset, images were aligned to MNI 152 space by affine transformation. The final images were cropped to size ($162\times192\times144$) and normalized by the maximum intensity of each volume.

\vspace{-0.5em}
\subsection{Experimental Setting}
\vspace{-0.5em}
The proposed method is evaluated on the atlas-based image registration task same as \cite{mok2020fast}. We compare our method with traditional optimization-based methods: SyN\cite{avants2008symmetric} and NiftyReg\cite{modat2010fast}, and state-of-the-art learning-based methods VoxelMorph(VM)\cite{balakrishnan2018unsupervised} and SYMNet\cite{mok2020fast}. For each dataset, we randomly sampled 20 scans as moving images and 3 scans as the atlases, resulting in 60 image pairs, and evaluate the results using all anatomical labels. To make a fair comparison, we also conduct instance-specific optimization for the learning-based methods and compare the optimization results. We also compare DNVF with two recent independently proposed methods, IDIR \cite{wolterink2021implicit} and NODEO \cite{wu2022nodeo} (released at roughly the same time as the preprint of this work) following NODEO's data setting. Both methods also use neural nets to model deformation but with some major differences: IDIR models deformation field instead of velocity field, while NODEO involves CNN.  

We follow the parameter setting of SyN in VoxelMorph with gradient step size 0.25 and gaussian parameter [0.9, 0.2]. Both SyN and NiftyReg use cross-correlation as the cost function. For learning-based methods, we use 86 and 250 images as the training data for Mindboggle and OASIS dataset respectively. Different iteration settings were used to analyze the running time and performance of optimization-based methods.

During the optimization of DNVF, ${\lambda_{1}}$ and ${\lambda_{2}}$ are empirically set to 100 and 0.1 for the local orientation consistency and the smoothness of deformation. The FCN of Cas-DNVF is trained with ${\lambda_{1}}$ and ${\lambda_{2}}$ being 10 and 4. The network parameters are optimized using the Adam algorithm with a learning rate of $1e^{-4}$. Our model is implemented using PyTorch and evaluated on a machine with a RTX 2080 Ti GPU and an Intel i7-7700K CPU.

\vspace{-0.5em}
\subsection{Evaluation Metric}
\vspace{-0.5em}
The goal of diffeomorphic image registration is to generate spatial correspondences between pairs of images while maintaining the topology. We evaluate the performance of registration methods using Dice Similarity Coefficient, Jacobian Determinant, and the Structural Similarity following \cite{dalca2018unsupervised,mok2020fast,mok2020large,zhang2021learning}. Dice Similarity Coefficient (\textbf{DSC}) measures the overlap between the segmentation of the fixed image and the warped segmentation of the moving image based on the deformation field. Negative Jacobian Determinant (\textbf{$\boldsymbol{|J_{<0}|}$}) represents local distortion in the neighborhood as discussed in Sec. \ref{sec:opt}. We report the ratio of ${|J_{<0}|}$ to evaluate to performance of topology preservation. Structural Similarity (\textbf{SSIM}) \cite{wang2004image} measures the similarity between fixed image and warped moving image by taking texture into account.

\vspace{-0.5em}
\subsection{Results Comparison and Discussion}
\vspace{-0.5em}

\input{tabs/tab2}
\vspace{-0.5em}
\paragraph{Accuracy} The evaluation results of our methods, compared to both traditional optimization-based and learning-based methods, are summarized in Table \ref{tab:results_comp}.
We report the results of SyN with iteration setting [600,600,300] and the results of NiftyReg with maximal level and iteration [5, 1000], which give better results than their default settings. All available anatomical masks are used in the evaluation. DNVF outperforms the traditional methods on Mindboggle and OASIS dataset in terms of DSC and SSIM, and still achieves the low ratio of ${|J_{<0}|}$. The evaluation results show the benefit of flexibility brought by neural velocity field compared with traditional methods. DNVF also outperforms the learning-based methods on two datasets by a large margin which demonstrates the capability of DNVF in modelling small local deformations while preserving the local orientation. 

Cas-DNVF further improves the performance of DNVF which illustrates the benefit of involving the initial deformation predicted by a pretrained FCN. However, the ratio of negative Jacobian determinant is slightly larger than DNVF, we argue that is because the composition of initial and residual deformation during the optimization introduces additional disturbance of deformation. We also compare the optimization result of learning-based methods with DNVF and Cas-DNVF. Specifically, for each pair of test images, we finetune the pretrained VoxelMorph and SYMNet. Both VoxelMorph and SYMNet generate better results after the optimization, but still underperform the DNVF and Cas-DNVF according to the DSC and SSIM.

\begin{figure}[ht]
\begin{center}
\includegraphics[width=\linewidth]{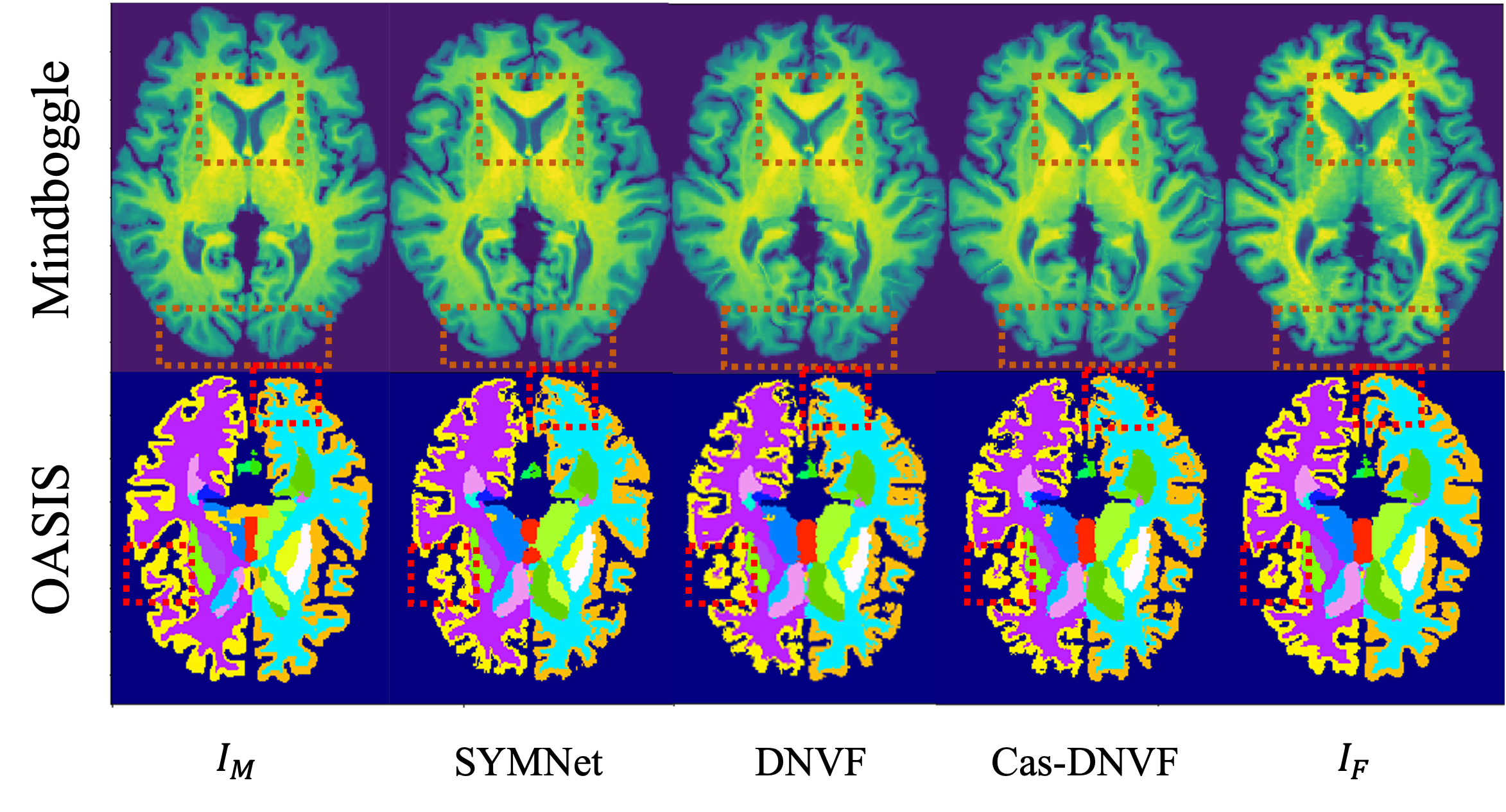}
\end{center}
\vspace{-1em}
   \caption{Visualization of registration results. First row compares the warped images from Mindboggle dataset and the second row shows the anatomical structures after the registration for OASIS dataset.}
\label{fig:vis}
\vspace{-1.5em}
\end{figure}

\paragraph{Visualization} Figure \ref{fig:vis} shows some visualization of registration results. Current SOTA method provides a reasonable matching between pair of images, however, some local small deformations are missing in both cortical and subcortical regions. Benefiting from the our neural velocity field representation, DNVF and Cas-DNVF are able to capture the local small deformations with better capability in fitting high-frequency content. 

\begin{figure*}[!htbp]
\begin{center}
\includegraphics[width=0.65\linewidth]{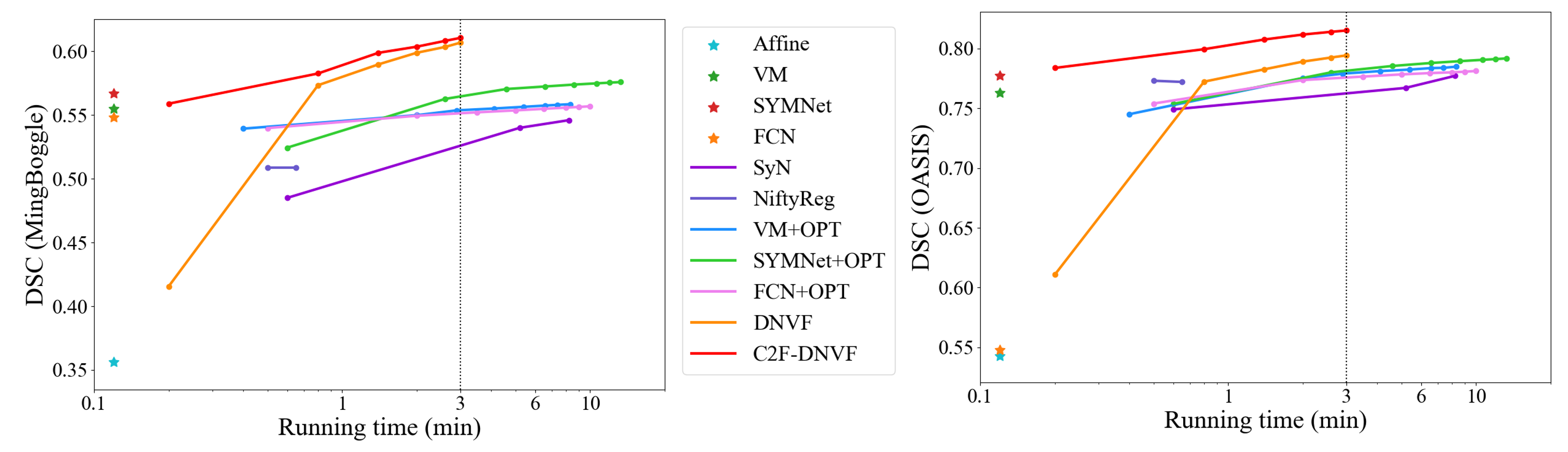}
\end{center}
\vspace{-1em}
   \caption{DSC versus Running time with different iteration setting.}
\label{fig:dsc_time}
\vspace{-1.5em}
\end{figure*}

\paragraph{Speed} Figure \ref{fig:dsc_time} compares the DSC versus running time for each method. The proposed DNVF outperforms the state-of-the-art traditional and DLIR methods with \textbf{48s} (80 iterations) optimization on the MindBoggle dataset and \textbf{72s} (120 iterations) optimization on the OASIS dataset. By combining the initial deformation predicted from our pre-trained FCN, Cas-NeVF achieves better performance in both matching accuracy and time efficiency. The maximum optimization step of DNVF and Cas-DNVF is set to 300 iterations (180s), but the performance will be further improved with more iterations according to the curve in Fig.\ref{fig:dsc_time}.

\vspace{-1em}
\begin{table} [H]
  \begin{center}
    {\small{
\setlength{\tabcolsep}{1mm}{
\begin{tabular}{cccccc}

\toprule
\multicolumn{1}{c}{Dataset}  & \multicolumn{2}{c}{OASIS} & \multicolumn{3}{c}{CANDI \cite{kennedy2012candishare}}   \\
\cmidrule(lr){2-3} \cmidrule(lr){4-6}
 \scriptsize{Model/Metric}    & \scriptsize{$\text{DSC}_\text{28}$ ($\uparrow$)} & \scriptsize{$|J_{<0}| (\downarrow$)} &  \scriptsize{$\text{DSC}_\text{32}$ ($\uparrow$)} & \scriptsize{$\text{DSC}_\text{28}$ ($\uparrow$)} & \scriptsize{$|J_{<0}| (\downarrow$)}\\
\midrule
IDIR  & 0.794 & 0.124\% & 0.774 & 0.811 & 0.113\% \\
NODEO & 0.779 & 0.03\% & 0.760 & 0.802 & \textbf{1.8×$\text{10}^\textbf{-7}$\%}\\
\midrule
DNVF & \textbf{0.815} & \textbf{0.0004\%} & \textbf{0.786} & \textbf{0.823} & 0.0003\%\\
\bottomrule
\end{tabular}
}}
}
\end{center}
\vspace{-0.5em}
\caption{Following NODEO, we use the same data split and evaluation metric. The DSC for OASIS is averaged over \textbf{28} structures excluding the very small ones and the DSC for CANDI is averaged over all \textbf{32} structures and \textbf{28} large structures as stated in NODEO's setting. We limit the optimization of DNVF to 100 iteration (60s) compared with NODEO (80s)}
\vspace{-1.5em}
\label{tab:1}
\end{table}

\paragraph{Table \ref{tab:1}} shows the evaluation results between DNVF and recent optimization-based methods. The comparison is conducted on OASIS and CANDI dataset \cite{kennedy2012candishare} which were used in NODEO's paper and we directly adopt the results as the implementation is not available. $|J_{<0}|$ reflects local distortion in deformation field. The diffeomorphism realized by velocity integration provides desired properties such as deformation smoothing and topology preservation for DNVF and NODEO. Therefore they have better performance in $|J_{<0}|$ than IDIR which doesn't realize diffeomorphism but models the displacement field. NODEO involves the convolutional layer and gaussian kernel to enhance the spatial interaction and reduce the ratio of ${|J_{<0}|}$. However, it provides the lowest matching accuracy in terms of DSC which is consistent with the result in Ablation study 2. Moreover, NODEO utilizes Neural ODE solver to do the integration requiring longer time to converge, which is not desired in real medical application. DNVF achieves the highest DSC score and lowest ratio of ${|J_{<0}|}$ which demonstrate the benefit of using MLP-based neural field and periodic sinusoidal activation functions to model the diffeomorphic deformation. Besides, DNVF only requires one-pass of the neural velocity field and utilizing scaling and squaring does not introduce additional learnable parameters, therefore it defines a simpler deformation space than NODEO and reduces the difficulty in finding the optimal solution via limited steps of optimization.

\vspace{-0.5em}
\subsection{Ablation Study}
\label{sec:aba}
\vspace{-0.5em}
In this section, we conduct ablation studies to measure the impact of components in our proposed method in terms of DSC which is averaged on all anatomical structures.
\vspace{-0.5em}

\paragraph{Activation function} In Sec. \ref{sec:4.1}, we state the benefit of using periodic sinusoidal activation to model small local deformation in diffeomorphic image registration. In this study, we conduct experiments on Mindboggle and OASIS datasets using classic ReLU activation with or without positional encoding (PE) as shown in Table \ref{tab:tab3}. Though the fixed PE performed well in \cite{mildenhall2020nerf} with the supervision of ground truth, it didn't help to capture the dynamic dense correspondence and deformation field in unsupervised image registration task. The results show that the nested sinusoidal activation function provides better capacity in capturing the correspondence and modeling deformation.
\vspace{-1em}
\begin{table} [H]
  \begin{center}
    {\small{
\begin{tabular}{lccc}
\toprule
Dataset & DNVF & MLP+ReLU & MLP+ReLU+PE \\
\midrule
Mindboggle  & \textbf{0.606} & 0.440 & 0.397\\
OASIS & \textbf{0.794} & 0.708 & 0.683\\
\bottomrule
\end{tabular}
}}
\end{center}
\vspace{-0.5em}
\caption{Results using sinusoidal and ReLU activation functions.}
\vspace{-1.5em}
\label{tab:tab3}
\end{table}
\vspace{-0.2em}
\paragraph{Velocity field representation} Table \ref{tab:tab4} shows the results with different velocity field representation: I) \textbf{Grid}: We use volumetric learnable parameter with size ($D\times H \times W \times 3$) to represent the entire velocity field and update this parameter via optimization with additional regularization term (TV) suggested by \cite{fridovich2022plenoxels}; II) \textbf{CNN}: Because the input to DNVF is a downsampled coordinate grid with shape ($\frac{D}{3}\times \frac{H}{3}\times \frac{W}{3}\times 3$), we replace the MLP with a 3D convolution neural network. The evaluation results demonstrate that the fourier mapping expressed by the MLP in (6) provides the high deformation representation power and matching accuracy. 
\vspace{-0.8em}
\begin{table} [H]
  \begin{center}
    {\small{
\begin{tabular}{lccccc}
\toprule
Dataset & DNVF & Grid & Conv+Sin & Conv+ReLU\\
\midrule
Mindboggle  & \textbf{0.606} & 0.572 & 0.515 & 0.480\\
OASIS & \textbf{0.794} & 0.755 & 0.739 & 0.731\\
\bottomrule
\end{tabular}
}}
\end{center}
\vspace{-0.5em}
\caption{Results with different velocity field representations.}
\vspace{-1.5em}
\label{tab:tab4}
\end{table}
\vspace{-0.5em}
\paragraph{Choice of pretrained FCN for Cas-DNVF}\label{aba:3} The proposed DNVF can also work with SOTA learning-based methods as shown in Table \ref{tab:tab5}. In this study, we replace the original FCN with VoxelMorph(VM) and SYMNet. The results demonstrate that the Cas-DNVF is a generalized framework and provide consistent performance improvement to different learning-based methods.
\vspace{-0.8em}
\begin{table} [H]
  \begin{center}
    {\small{
\setlength{\tabcolsep}{1.5mm}{
\begin{tabular}{lccc}
\toprule
Dataset & FCN+DNVF & VM+DNVF & SYMNet+DNVF  \\
\midrule
Mindboggle  & \textbf{0.612} & 0.609 & 0.606 \\
OASIS & 0.815 & \textbf{0.816} & 0.812\\
\bottomrule
\end{tabular}
}
}}
\end{center}
\vspace{-0.5em}
\caption{Results with different pretrained FCN in Cas-DNVF.}
\vspace{-0.5em}
\label{tab:tab5}
\end{table}
\vspace{-1.5em}

\section{Conclusion}
\vspace{-0.5em}
In this paper, we propose a neural field model to represent the continuous velocity field and model deformation for solving diffeomorphic image registration. The validation experiments demonstrate the significant advantages brought by proposed methods. 
The proposed DNVF and Cas-DNVF methods offer a new framework for classical image registration problem. 
However, our current methods still have some limitations. 
First, the model has a relatively large memory footprint due to scaling and squaring component. A future direction is to improve the derivation of deformation field from velocity field. 
%
Second, the two stages of Cas-DNVF is decoupled.  A future improvement is to integrate them and train the learning model in an end-to-end manner.

{\small
\bibliographystyle{ieee_fullname}
\bibliography{egbib}
}

\end{document}

%% file: tabs/tab2.tex
\begin{table*}[ht]
\centering
\begin{center}
\begin{tabular}{lcccccccc}
\toprule
& \multicolumn{1}{c}{Dataset}  & \multicolumn{3}{c}{Mindboggle \cite{marcus2007open}} & \multicolumn{3}{c}{OASIS \cite{klein2012101}}   \\
\cmidrule(lr){3-5} \cmidrule(lr){6-8}
Category                                & \scriptsize{Model/Metric}    & \scriptsize{DSC($\uparrow$)} & \scriptsize{$|J_{<0}| (\downarrow$)} & \scriptsize{SSIM($\uparrow$)} & \scriptsize{DSC($\uparrow$)} & \scriptsize{$|J_{<0}| (\downarrow$)} & \scriptsize{SSIM($\uparrow$)}      \\
\midrule
\multirow{2}{*}{Traditional Methods}
                                        & \footnotesize{SyN \cite{avants2008symmetric}}        & 0.548    & \textbf{$\approx$0}    & 0.9097  & 0.777  & \textbf{$\approx$0} & 0.9149        \\
                                        & \footnotesize{NiftyReg \cite{modat2010fast}}        & 0.509    & \textbf{$\approx$0}    & 0.8650  & 0.773  & \textbf{$\approx$0} & 0.8916 \\
\midrule
\multirow{3}{*}{Learning-based Methods}         & \footnotesize{VoxelMorph \cite{balakrishnan2018unsupervised}}        & 0.555    & 0.076\%    & 0.9145  & 0.763  & 0.072\%  & 0.9082     \\
                                        & \footnotesize{SYMNet \cite{mok2020fast}}        & 0.567    & 0.0023\%    & 0.9191  & 0.777  &0.0022\% & 0.9247 \\
                                        & \footnotesize{FCN}        & 0.548    & \textbf{$\approx$0}    & 0.9017  & 0.765  & \textbf{$\approx$0} & 0.9091 \\
\midrule
Instance-specific         & \footnotesize{VoxelMorph+OPT}        & 0.558    & 0.072\%    & 0.9186  & 0.784  &0.051\% & 0.9141     \\
Optimization of                                        & \footnotesize{SYMNet+OPT}        & 0.575    & 0.0010\%    & 0.9271  & 0.791  &0.0012\% & 0.9292         \\
Learning-based Methods                                        & \footnotesize{FCN+OPT}        & 0.556    & \textbf{$\approx$0}    & 0.9072  & 0.781  & \textbf{$\approx$0} & 0.9137         \\
\midrule
\multirow{2}{*}{Our Methods}         & \footnotesize{DNVF}        & \textbf{0.606}    & 0.0013\%    & \textbf{0.9496}  & \textbf{0.794}  &0.0015\% & \textbf{0.9512}     \\
                                        & \footnotesize{Cas-DNVF}        & \textbf{0.612}    & 0.0031\%    & \textbf{0.9531}  & \textbf{0.815} & 0.0036\%  & \textbf{0.9532}         \\

\bottomrule

\end{tabular}
\end{center}
\vspace{-0.5em}
\caption{
The registration methods are evaluated using \textbf{all} available anatomical structures (\textbf{31} for Mindboggle and \textbf{35} for OASIS) in atlas-based fashion. FCN is the model used in C2F-DNVF which provides better topology preservation compared with other learning-based methods. Table \ref{tab:tab5} shows the results of replacing FCN with other SOTA methods in C2F-DNVF. For instance-specific optimization of learning-based methods, we finetune the pretrained model for each given pair of test images. 
}
\label{tab:results_comp}
\vspace{-1.7em}
\end{table*}

%% file: main.bbl
\begin{thebibliography}{10}\itemsep=-1pt

\bibitem{arsigny2006log}
Vincent Arsigny, Olivier Commowick, Xavier Pennec, and Nicholas Ayache.
\newblock A log-euclidean framework for statistics on diffeomorphisms.
\newblock In {\em International Conference on Medical Image Computing and
  Computer-Assisted Intervention}, pages 924--931. Springer, 2006.

\bibitem{ashburner2000voxel}
John Ashburner and Karl~J Friston.
\newblock Voxel-based morphometry—the methods.
\newblock {\em Neuroimage}, 11(6):805--821, 2000.

\bibitem{avants2008symmetric}
Brian~B Avants, Charles~L Epstein, Murray Grossman, and James~C Gee.
\newblock Symmetric diffeomorphic image registration with cross-correlation:
  evaluating automated labeling of elderly and neurodegenerative brain.
\newblock {\em Medical image analysis}, 12(1):26--41, 2008.

\bibitem{bajcsy1989multiresolution}
Ruzena Bajcsy and Stane Kova{\v{c}}i{\v{c}}.
\newblock Multiresolution elastic matching.
\newblock {\em Computer vision, graphics, and image processing}, 46(1):1--21,
  1989.

\bibitem{balakrishnan2018unsupervised}
Guha Balakrishnan, Amy Zhao, Mert~R Sabuncu, John Guttag, and Adrian~V Dalca.
\newblock An unsupervised learning model for deformable medical image
  registration.
\newblock In {\em Proceedings of the IEEE conference on computer vision and
  pattern recognition}, pages 9252--9260, 2018.

\bibitem{basri2020frequency}
Ronen Basri, Meirav Galun, Amnon Geifman, David Jacobs, Yoni Kasten, and Shira
  Kritchman.
\newblock Frequency bias in neural networks for input of non-uniform density.
\newblock In {\em International Conference on Machine Learning}, pages
  685--694. PMLR, 2020.

\bibitem{beg2005computing}
M~Faisal Beg, Michael~I Miller, Alain Trouv{\'e}, and Laurent Younes.
\newblock Computing large deformation metric mappings via geodesic flows of
  diffeomorphisms.
\newblock {\em International journal of computer vision}, 61(2):139--157, 2005.

\bibitem{benbarka2022seeing}
Nuri Benbarka, Timon H{\"o}fer, Andreas Zell, et~al.
\newblock Seeing implicit neural representations as fourier series.
\newblock In {\em Proceedings of the IEEE/CVF Winter Conference on Applications
  of Computer Vision}, pages 2041--2050, 2022.

\bibitem{chen2021vit}
Junyu Chen, Yufan He, Eric~C Frey, Ye Li, and Yong Du.
\newblock Vit-v-net: Vision transformer for unsupervised volumetric medical
  image registration.
\newblock {\em arXiv preprint arXiv:2104.06468}, 2021.

\bibitem{chen2021transunet}
Jieneng Chen, Yongyi Lu, Qihang Yu, Xiangde Luo, Ehsan Adeli, Yan Wang, Le Lu,
  Alan~L Yuille, and Yuyin Zhou.
\newblock Transunet: Transformers make strong encoders for medical image
  segmentation.
\newblock {\em arXiv preprint arXiv:2102.04306}, 2021.

\bibitem{chen2021deep}
Xuming Chen, Shanlin Sun, Narisu Bai, Kun Han, Qianqian Liu, Shengyu Yao, Hao
  Tang, Chupeng Zhang, Zhipeng Lu, Qian Huang, et~al.
\newblock A deep learning-based auto-segmentation system for organs-at-risk on
  whole-body computed tomography images for radiation therapy.
\newblock {\em Radiotherapy and Oncology}, 160:175--184, 2021.

\bibitem{christensen1996deformable}
Gary~E Christensen, Richard~D Rabbitt, and Michael~I Miller.
\newblock Deformable templates using large deformation kinematics.
\newblock {\em IEEE transactions on image processing}, 5(10):1435--1447, 1996.

\bibitem{dalca2018unsupervised}
Adrian~V Dalca, Guha Balakrishnan, John Guttag, and Mert~R Sabuncu.
\newblock Unsupervised learning for fast probabilistic diffeomorphic
  registration.
\newblock In {\em International Conference on Medical Image Computing and
  Computer-Assisted Intervention}, pages 729--738. Springer, 2018.

\bibitem{dupuis1998variational}
Paul Dupuis, Ulf Grenander, and Michael~I Miller.
\newblock Variational problems on flows of diffeomorphisms for image matching.
\newblock {\em Quarterly of applied mathematics}, pages 587--600, 1998.

\bibitem{fridovich2022plenoxels}
Sara Fridovich-Keil, Alex Yu, Matthew Tancik, Qinhong Chen, Benjamin Recht, and
  Angjoo Kanazawa.
\newblock Plenoxels: Radiance fields without neural networks.
\newblock In {\em Proceedings of the IEEE/CVF Conference on Computer Vision and
  Pattern Recognition}, pages 5501--5510, 2022.

\bibitem{hu2021end}
Jing Hu, Ziwei Luo, Xin Wang, Shanhui Sun, Youbing Yin, Kunlin Cao, Qi Song,
  Siwei Lyu, and Xi Wu.
\newblock End-to-end multimodal image registration via reinforcement learning.
\newblock {\em Medical Image Analysis}, 68:101878, 2021.

\bibitem{incoronato2017radiogenomic}
Mariarosaria Incoronato, Marco Aiello, Teresa Infante, Carlo Cavaliere,
  Anna~Maria Grimaldi, Peppino Mirabelli, Serena Monti, and Marco Salvatore.
\newblock Radiogenomic analysis of oncological data: a technical survey.
\newblock {\em International journal of molecular sciences}, 18(4):805, 2017.

\bibitem{kennedy2012candishare}
David~N Kennedy, Christian Haselgrove, Steven~M Hodge, Pallavi~S Rane, Nikos
  Makris, and Jean~A Frazier.
\newblock Candishare: a resource for pediatric neuroimaging data, 2012.

\bibitem{klein2012101}
Arno Klein and Jason Tourville.
\newblock 101 labeled brain images and a consistent human cortical labeling
  protocol.
\newblock {\em Frontiers in neuroscience}, 6:171, 2012.

\bibitem{li2021neural}
Zhengqi Li, Simon Niklaus, Noah Snavely, and Oliver Wang.
\newblock Neural scene flow fields for space-time view synthesis of dynamic
  scenes.
\newblock In {\em Proceedings of the IEEE/CVF Conference on Computer Vision and
  Pattern Recognition}, pages 6498--6508, 2021.

\bibitem{marcus2007open}
Daniel~S Marcus, Tracy~H Wang, Jamie Parker, John~G Csernansky, John~C Morris,
  and Randy~L Buckner.
\newblock Open access series of imaging studies (oasis): cross-sectional mri
  data in young, middle aged, nondemented, and demented older adults.
\newblock {\em Journal of cognitive neuroscience}, 19(9):1498--1507, 2007.

\bibitem{mildenhall2020nerf}
Ben Mildenhall, Pratul~P Srinivasan, Matthew Tancik, Jonathan~T Barron, Ravi
  Ramamoorthi, and Ren Ng.
\newblock Nerf: Representing scenes as neural radiance fields for view
  synthesis.
\newblock In {\em European conference on computer vision}, pages 405--421.
  Springer, 2020.

\bibitem{modat2010fast}
Marc Modat, Gerard~R Ridgway, Zeike~A Taylor, Manja Lehmann, Josephine Barnes,
  David~J Hawkes, Nick~C Fox, and S{\'e}bastien Ourselin.
\newblock Fast free-form deformation using graphics processing units.
\newblock {\em Computer methods and programs in biomedicine}, 98(3):278--284,
  2010.

\bibitem{mok2020fast}
Tony~CW Mok and Albert Chung.
\newblock Fast symmetric diffeomorphic image registration with convolutional
  neural networks.
\newblock In {\em Proceedings of the IEEE/CVF conference on computer vision and
  pattern recognition}, pages 4644--4653, 2020.

\bibitem{mok2020large}
Tony~CW Mok and Albert Chung.
\newblock Large deformation diffeomorphic image registration with laplacian
  pyramid networks.
\newblock In {\em International Conference on Medical Image Computing and
  Computer-Assisted Intervention}, pages 211--221. Springer, 2020.

\bibitem{mok2021conditional}
Tony~CW Mok and Albert Chung.
\newblock Conditional deformable image registration with convolutional neural
  network.
\newblock In {\em International Conference on Medical Image Computing and
  Computer-Assisted Intervention}, pages 35--45. Springer, 2021.

\bibitem{mok2022affine}
Tony~CW Mok and Albert Chung.
\newblock Affine medical image registration with coarse-to-fine vision
  transformer.
\newblock In {\em Proceedings of the IEEE/CVF Conference on Computer Vision and
  Pattern Recognition}, pages 20835--20844, 2022.

\bibitem{niemeyer2019occupancy}
Michael Niemeyer, Lars Mescheder, Michael Oechsle, and Andreas Geiger.
\newblock Occupancy flow: 4d reconstruction by learning particle dynamics.
\newblock In {\em Proceedings of the IEEE/CVF international conference on
  computer vision}, pages 5379--5389, 2019.

\bibitem{park2019deepsdf}
Jeong~Joon Park, Peter Florence, Julian Straub, Richard Newcombe, and Steven
  Lovegrove.
\newblock Deepsdf: Learning continuous signed distance functions for shape
  representation.
\newblock In {\em Proceedings of the IEEE/CVF conference on computer vision and
  pattern recognition}, pages 165--174, 2019.

\bibitem{pumarola2021d}
Albert Pumarola, Enric Corona, Gerard Pons-Moll, and Francesc Moreno-Noguer.
\newblock D-nerf: Neural radiance fields for dynamic scenes.
\newblock In {\em Proceedings of the IEEE/CVF Conference on Computer Vision and
  Pattern Recognition}, pages 10318--10327, 2021.

\bibitem{rahaman2019spectral}
Nasim Rahaman, Aristide Baratin, Devansh Arpit, Felix Draxler, Min Lin, Fred
  Hamprecht, Yoshua Bengio, and Aaron Courville.
\newblock On the spectral bias of neural networks.
\newblock In {\em International Conference on Machine Learning}, pages
  5301--5310. PMLR, 2019.

\bibitem{risholm2011multimodal}
Petter Risholm, Alexandra~J Golby, and William Wells.
\newblock Multimodal image registration for preoperative planning and
  image-guided neurosurgical procedures.
\newblock {\em Neurosurgery Clinics}, 22(2):197--206, 2011.

\bibitem{ronneberger2015u}
Olaf Ronneberger, Philipp Fischer, and Thomas Brox.
\newblock U-net: Convolutional networks for biomedical image segmentation.
\newblock In {\em International Conference on Medical image computing and
  computer-assisted intervention}, pages 234--241. Springer, 2015.

\bibitem{rueckert1999nonrigid}
Daniel Rueckert, Luke~I Sonoda, Carmel Hayes, Derek~LG Hill, Martin~O Leach,
  and David~J Hawkes.
\newblock Nonrigid registration using free-form deformations: application to
  breast mr images.
\newblock {\em IEEE transactions on medical imaging}, 18(8):712--721, 1999.

\bibitem{sheikhjafari2018unsupervised}
Ameneh Sheikhjafari, Michelle Noga, Kumaradevan Punithakumar, and Nilanjan Ray.
\newblock Unsupervised deformable image registration with fully connected
  generative neural network.
\newblock 2018.

\bibitem{shen2002hammer}
Dinggang Shen and Christos Davatzikos.
\newblock Hammer: hierarchical attribute matching mechanism for elastic
  registration.
\newblock {\em IEEE transactions on medical imaging}, 21(11):1421--1439, 2002.

\bibitem{shen2021accurate}
Zhengyang Shen, Jean Feydy, Peirong Liu, Ariel~H Curiale, Ruben San
  Jose~Estepar, Raul San Jose~Estepar, and Marc Niethammer.
\newblock Accurate point cloud registration with robust optimal transport.
\newblock {\em Advances in Neural Information Processing Systems},
  34:5373--5389, 2021.

\bibitem{shen2019networks}
Zhengyang Shen, Xu Han, Zhenlin Xu, and Marc Niethammer.
\newblock Networks for joint affine and non-parametric image registration.
\newblock In {\em Proceedings of the IEEE/CVF Conference on Computer Vision and
  Pattern Recognition}, pages 4224--4233, 2019.

\bibitem{shen2019region}
Zhengyang Shen, Fran{\c{c}}ois-Xavier Vialard, and Marc Niethammer.
\newblock Region-specific diffeomorphic metric mapping.
\newblock {\em Advances in Neural Information Processing Systems}, 32, 2019.

\bibitem{shi2022xmorpher}
Jiacheng Shi, Yuting He, Youyong Kong, Jean-Louis Coatrieux, Huazhong Shu,
  Guanyu Yang, and Shuo Li.
\newblock Xmorpher: Full transformer for deformable medical image registration
  via cross attention.
\newblock {\em arXiv preprint arXiv:2206.07349}, 2022.

\bibitem{sitzmann2020implicit}
Vincent Sitzmann, Julien Martel, Alexander Bergman, David Lindell, and Gordon
  Wetzstein.
\newblock Implicit neural representations with periodic activation functions.
\newblock {\em Advances in Neural Information Processing Systems},
  33:7462--7473, 2020.

\bibitem{sotiras2013deformable}
Aristeidis Sotiras, Christos Davatzikos, and Nikos Paragios.
\newblock Deformable medical image registration: A survey.
\newblock {\em IEEE transactions on medical imaging}, 32(7):1153--1190, 2013.

\bibitem{sun2022topology}
Shanlin Sun, Kun Han, Deying Kong, Hao Tang, Xiangyi Yan, and Xiaohui Xie.
\newblock Topology-preserving shape reconstruction and registration via neural
  diffeomorphic flow.
\newblock In {\em Proceedings of the IEEE/CVF Conference on Computer Vision and
  Pattern Recognition}, pages 20845--20855, 2022.

\bibitem{tancik2020fourier}
Matthew Tancik, Pratul Srinivasan, Ben Mildenhall, Sara Fridovich-Keil, Nithin
  Raghavan, Utkarsh Singhal, Ravi Ramamoorthi, Jonathan Barron, and Ren Ng.
\newblock Fourier features let networks learn high frequency functions in low
  dimensional domains.
\newblock {\em Advances in Neural Information Processing Systems},
  33:7537--7547, 2020.

\bibitem{tang2021spatial}
Hao Tang, Xingwei Liu, Kun Han, Xiaohui Xie, Xuming Chen, Huang Qian, Yong Liu,
  Shanlin Sun, and Narisu Bai.
\newblock Spatial context-aware self-attention model for multi-organ
  segmentation.
\newblock In {\em Proceedings of the IEEE/CVF Winter Conference on Applications
  of Computer Vision}, pages 939--949, 2021.

\bibitem{tang2021recurrent}
Hao Tang, Xingwei Liu, Shanlin Sun, Xiangyi Yan, and Xiaohui Xie.
\newblock Recurrent mask refinement for few-shot medical image segmentation.
\newblock In {\em Proceedings of the IEEE/CVF International Conference on
  Computer Vision}, pages 3918--3928, 2021.

\bibitem{thirion1998image}
J-P Thirion.
\newblock Image matching as a diffusion process: an analogy with maxwell's
  demons.
\newblock {\em Medical image analysis}, 2(3):243--260, 1998.

\bibitem{wang2004image}
Zhou Wang, Alan~C Bovik, Hamid~R Sheikh, and Eero~P Simoncelli.
\newblock Image quality assessment: from error visibility to structural
  similarity.
\newblock {\em IEEE transactions on image processing}, 13(4):600--612, 2004.

\bibitem{wolterink2021implicit}
Jelmer~M Wolterink, Jesse~C Zwienenberg, and Christoph Brune.
\newblock Implicit neural representations for deformable image registration.
\newblock In {\em Medical Imaging with Deep Learning}, 2021.

\bibitem{wu2021irem}
Qing Wu, Yuwei Li, Lan Xu, Ruiming Feng, Hongjiang Wei, Qing Yang, Boliang Yu,
  Xiaozhao Liu, Jingyi Yu, and Yuyao Zhang.
\newblock Irem: High-resolution magnetic resonance image reconstruction via
  implicit neural representation.
\newblock In {\em International Conference on Medical Image Computing and
  Computer-Assisted Intervention}, pages 65--74. Springer, 2021.

\bibitem{wu2022nodeo}
Yifan Wu, Tom~Z Jiahao, Jiancong Wang, Paul~A Yushkevich, M~Ani Hsieh, and
  James~C Gee.
\newblock Nodeo: A neural ordinary differential equation based optimization
  framework for deformable image registration.
\newblock In {\em Proceedings of the IEEE/CVF Conference on Computer Vision and
  Pattern Recognition}, pages 20804--20813, 2022.

\bibitem{xie2022neural}
Yiheng Xie, Towaki Takikawa, Shunsuke Saito, Or Litany, Shiqin Yan, Numair
  Khan, Federico Tombari, James Tompkin, Vincent Sitzmann, and Srinath Sridhar.
\newblock Neural fields in visual computing and beyond.
\newblock In {\em Computer Graphics Forum}, volume~41, pages 641--676. Wiley
  Online Library, 2022.

\bibitem{xu2021multi}
Junshen Xu, Eric~Z Chen, Xiao Chen, Terrence Chen, and Shanhui Sun.
\newblock Multi-scale neural odes for 3d medical image registration.
\newblock In {\em International Conference on Medical Image Computing and
  Computer-Assisted Intervention}, pages 213--223. Springer, 2021.

\bibitem{xu2019deepatlas}
Zhenlin Xu and Marc Niethammer.
\newblock Deepatlas: Joint semi-supervised learning of image registration and
  segmentation.
\newblock In {\em International Conference on Medical Image Computing and
  Computer-Assisted Intervention}, pages 420--429. Springer, 2019.

\bibitem{yan2022after}
Xiangyi Yan, Hao Tang, Shanlin Sun, Haoyu Ma, Deying Kong, and Xiaohui Xie.
\newblock After-unet: Axial fusion transformer unet for medical image
  segmentation.
\newblock In {\em Proceedings of the IEEE/CVF Winter Conference on Applications
  of Computer Vision}, pages 3971--3981, 2022.

\bibitem{you2022mine}
Chenyu You, Weicheng Dai, Fenglin Liu, Haoran Su, Xiaoran Zhang, Lawrence
  Staib, and James~S Duncan.
\newblock Mine your own anatomy: Revisiting medical image segmentation with
  extremely limited labels.
\newblock {\em arXiv preprint arXiv:2209.13476}, 2022.

\bibitem{you2022bootstrapping}
Chenyu You, Weicheng Dai, Lawrence Staib, and James~S Duncan.
\newblock Bootstrapping semi-supervised medical image segmentation with
  anatomical-aware contrastive distillation.
\newblock {\em arXiv preprint arXiv:2206.02307}, 2022.

\bibitem{you2022incremental}
Chenyu You, Jinlin Xiang, Kun Su, Xiaoran Zhang, Siyuan Dong, John Onofrey,
  Lawrence Staib, and James~S Duncan.
\newblock Incremental learning meets transfer learning: Application to
  multi-site prostate mri segmentation.
\newblock {\em arXiv preprint arXiv:2206.01369}, 2022.

\bibitem{you2022class}
Chenyu You, Ruihan Zhao, Fenglin Liu, Sandeep Chinchali, Ufuk Topcu, Lawrence
  Staib, and James~S Duncan.
\newblock Class-aware generative adversarial transformers for medical image
  segmentation.
\newblock {\em arXiv preprint arXiv:2201.10737}, 2022.

\bibitem{zhang2017frequency}
Miaomiao Zhang, Ruizhi Liao, Adrian~V Dalca, Esra~A Turk, Jie Luo, P~Ellen
  Grant, and Polina Golland.
\newblock Frequency diffeomorphisms for efficient image registration.
\newblock In {\em International conference on information processing in medical
  imaging}, pages 559--570. Springer, 2017.

\bibitem{zhang2022learning}
Xiaoran Zhang, Chenyu You, Shawn Ahn, Juntang Zhuang, Lawrence Staib, and James
  Duncan.
\newblock Learning correspondences of cardiac motion from images using
  biomechanics-informed modeling.
\newblock {\em arXiv preprint arXiv:2209.00726}, 2022.

\bibitem{zhang2021learning}
Yungeng Zhang, Yuru Pei, and Hongbin Zha.
\newblock Learning dual transformer network for diffeomorphic registration.
\newblock In {\em International Conference on Medical Image Computing and
  Computer-Assisted Intervention}, pages 129--138. Springer, 2021.

\bibitem{zhao2019recursive}
Shengyu Zhao, Yue Dong, Eric~I Chang, Yan Xu, et~al.
\newblock Recursive cascaded networks for unsupervised medical image
  registration.
\newblock In {\em Proceedings of the IEEE/CVF international conference on
  computer vision}, pages 10600--10610, 2019.

\end{thebibliography}
